\RequirePackage{fix-cm}
\documentclass[smallcondensed]{svjour3}     
\smartqed  
\usepackage{graphicx}
   \usepackage{lscape}
   \usepackage{natbib}
\usepackage{graphicx}
  \usepackage{lscape}
   \usepackage{natbib}
\begin{document}

\begin{center}
{\Large Locally Weighted Learning for Naive Bayes Classifier}

\vspace{0.6cm}
Kim-Hung Li

Department of Statistics, The Chinese University of Hong Kong

\vspace{0.3cm}
Cheuk Ting Li

Department of Electrical Engineering, Stanford University

\end{center}









\vspace{0.5cm}
\begin{abstract}
As a consequence of the  
strong and usually violated conditional independence assumption (CIA) of naive Bayes (NB) classifier, the performance of NB becomes less and less favorable compared to sophisticated classifiers when the sample size increases.  
We learn from this phenomenon that when the size of the training data is large, we should either relax the assumption or apply NB to a ``reduced'' data set, say for example use NB as a local model.  The latter approach trades the ignored information for the robustness to the model assumption.  In this paper, we consider using NB as a model for
locally weighted data.  
A special weighting function is designed so that if CIA holds for the unweighted data, it also holds for the weighted data.
The new method is intuitive and capable of handling class imbalance.
It is theoretically more sound than the locally weighted learners of naive Bayes that
base classification only on the $k$ nearest neighbors.  Empirical study
shows that the new method with appropriate choice of parameter outperforms 
seven existing classifiers of similar nature.  

\vspace{0.5cm}
\end{abstract}

\section{Introduction}
\label{intro}
Naive Bayes (NB) classifier is well-known for its simplicity,
computational efficiency, and competitive performance.  
Let $X = (X_{1},\ldots,X_{m})$
be a vector of $m$ feature variables (also known as attributes) and $Y$ be a class variable. 
NB makes a bold and usually violated assumption that $X_{1},\ldots,X_{m}$ are independent
given $Y$. This is the famous conditional independence assumption
(CIA).  In this paper, all feature variables are assumed to be categorical.  
CIA implies that 
\begin{eqnarray*}
& & \Pr(Y=y|X_{1}=x_{1},\ldots,X_{m}=x_{m}) \\
& \propto & \Pr(Y=y)\prod_{i=1}^{m}\Pr(X_{i}=x_{i}|Y=y).
\end{eqnarray*}
We call a realization
of $X$ an instance. An instance
can be labeled or unlabeled depending on whether the corresponding
value of $Y$ is observed or not. We have a training data set $\mathcal{T}$
comprising labeled instances which are random samples of $(X,Y)$.
We call any labeled instance in $\mathcal{T}$ a training instance. 
A test instance is an unlabeled instance whose $Y$-value we want to estimate.
For any test instance 
with $X = x^* = (x_{1}^{*},\ldots,x_{m}^{*})$, NB estimate of $Y$, denoted as $\hat{y}^{*}$,
is
\[
\hat{y}^{*}=\arg \max_{y}\widehat{\Pr}(Y=y)\prod_{i=1}^{m}\widehat{\Pr}(X_{i}=x_{i}^{*}|Y=y),
\]
where $\widehat{\Pr}(.)$ and $\widehat{\Pr}(.|.)$ are estimates
of the corresponding probabilities. Common probability estimator is
the relative frequency or its Laplace's modification \citep{cestnik1990}. 

An obvious weakness of NB is that the CIA is unrealistic in most real
applications. Although NB achieves quite good classification accuracy even 
when the CIA is violated by a wide margin,
CIA does have adverse effect on the asymptotic behavior of the classifier:
When we observe more and more data,
NB in general does not converge to the Bayesian classification rule for the full multinomial model. This characteristic of NB is confirmed by the empirical
results which show that when more data becomes available, the correct
classification rate of NB does not scale up \citep{kohavi1996},
and NB will eventually be overtaken by classifiers which make weaker
assumptions. 

Researches have been done to remedy this defect of NB.  They are developed along two directions.  One is to extend the NB model to a larger Bayesian network model.  The other modifies the training set to better suit the CIA.
An early attempt in the first direction is the tree augmented naive Bayes (TAN) model 
\citep{friedman1996,friedman1997b}
which embeds a tree topology on the Bayesian network of $(X,Y)$.  The averaged one-dependence estimator (AODE)
\citep{webb2005} is another method.  It uses the average of superparent-one-dependence estimators for classification.  \citet{cerquides2005} introduced a WAODE method
which extends AODE by replacing the simple average by a weighted average.  \citet{zhang2005} proposed
a hidden naive Bayes (HNB) classifier on which one hidden parent is added
to each attribute. 

In this paper, the second direction is taken.  We modify the training data by assigning each training instance a weight which depends on how close it is to the test instance.  NB is then fitted to the locally weighted data.  The unfavorable impact of CIA is lessened as focus is laid on a small neighborhood of the test instance.  

The remaining part of the paper is organized as follows. 
We consider how we can apply NB in the presence of weights in Section \ref{sec:Naive-Bayes-with-weights}.
The new classifier, which we call a lazy cell-weighted naive Bayes method, is
proposed in Section \ref{sec:Lazy-cell-weighting}. Empirical study
is performed in Section \ref{sec:Empirical-comparison} comparing
the new method with seven commonly used classifiers of similar nature.
The new method with an appropriate choice of parameter is found outperforming methods considered in the study. 
The last section concludes the paper.

\section{Naive Bayes with weights}
\label{sec:Naive-Bayes-with-weights}
Let $\Pr(x,y)$ and $w(x,y)$ be the probability and the (nonnegative) weight
of $(X=x,Y=y)$ respectively.
Define a random vector $(X,Y)$ with weight $w(x,y)$
to be a random vector with the same support as $(X,Y)$,
but with the joint probability 
\[
\mbox{Pr}_{w}(X=x, Y=y)\propto w(x,y)\Pr(X=x,Y=y).
\]
This distribution is well-defined as far as $w(x,y)\Pr(X=x,Y=y) > 0$
for at least one $(x,y)$. 

If all weights
are equal, the weighted random
vector reduces to the unweighted one. As weight is assigned to 
cell, we call it cell weight. To avoid confusion, we denote a random
vector with weight by $(X,Y)_{w}$. It 
satisfies the CIA if for any $x = (x_1, \ldots, x_m)$
\begin{equation}
\mbox{Pr}_{w}(Y=y|X=x)\propto\mbox{Pr}_{w}(Y=y)\prod_{i=1}^{m}\mbox{Pr}_{w}(X_i = x_{i}|Y=y).
\end{equation}
Under CIA, our estimate of $Y$ for a test instance with $X = x^* = (x_1^*, \ldots, x_m^*)$ is
\[
\hat{y}_{w}^{*}=\arg \max_{y}\widehat{\Pr}_{w}(Y=y)\prod_{i=1}^{m}\widehat{\Pr}_{w}(X_{i}=x_{i}^{*}|Y=y),
\]
where $\widehat{\Pr}_{w}$ denotes an estimate of $\mbox{Pr}_{w}$.
The weighted relative frequency estimators of probabilities are 
\[
\widehat{\Pr}_{w}(Y=y)=\frac{\sum_{x}f(x,y)w(x,y)}{\sum_{x,k}f(x,k)w(x,k)}
\]
 and
\begin{eqnarray*}
& & \widehat{\Pr}_{w}(X_{i}=x_{i}^{*}|Y=y) \\
& = &\frac{\sum_{x_{1},\ldots x_{i-1},x_{i+1},\ldots,x_{m}}f((x_{1},...,x_{i}^{*},...,x_{m}),y)w((x_{1},...,x_{i}^{*},...,x_{m}),y)}{\sum_{x}f(x,y)w(x,y)},
\end{eqnarray*}
where $f(x,y)$ is the observed frequency
of $X=x$ and $Y=y$ in $\mathcal{T}$. 

\section{Lazy cell-weighted naive Bayes}
\label{sec:Lazy-cell-weighting}

Locally weighted classifier is determined by a weighting function and a model.  The former assigns a nonnegative weight to each training instance so that instances ``closer'' to the test instance, $X= x^*$, have larger weights.  The model is then fitted to the weighted training data.  Classification is made basing on the estimated posterior probability of $Y$ under the fitted model.

\vspace{0.3cm}
\noindent
{\bf Definition.} Let $w_{x^*}(x,y)$ be a local weighting function for a test instance $x^*$.  It is called {\it compatible} to a model if 

\begin{description}
\item[]
(a) If the model holds for the unweighted data, it also holds for the weighted data, and

\item[]
(b) $w_{x^*}(x^*,y)$ does not depend on $y$.  

\end{description}

\vspace{0.3cm}
\noindent
Compatibility is fundamental for a weighting function and a model to form a reasonable locally weighted classifier if the aim of the weighting is to alleviate the possible failure of the model in a large region, rather than an intentional modification of the model.  In other words, the weighting works as a means to weaken the sensitivity of the classifier to the model instead of as a technique to introduce a new model.  Note that Condition (b) allows $w_{x^*}(x,y)$ to depend on $y$ when $x \neq x^*$.  This Condition ensures that 
\[
\mbox{Pr}_{w_{x^*}}(Y=y |X= x^*) = \Pr(Y=y |X=x^*).
\]
Therefore, estimator of the former using the weighted data is just an alternative estimator of the latter.  Hopefully, this new estimator is more robust than the original estimator to the model assumption.

The use of NB as a local model is not new.  A successful example is the hybrid classifier with decision tree.  A decision tree is built.  For each leaf, a NB is fitted to the data associated with that leaf \citep{kohavi1996, gama2003}.  The weighting function of this approach is compatible to NB when the feature space is partitioned by axis-parallel surfaces.  Another approach utilizes only the $k$ nearest neighbors of the test instance 
\citep{frank2003,jiang2005}.  Unfortunately, the weighting function is not compatible to NB
because of the conflict between CIA and the zero weight.
In this section, we propose a locally weighting function compatible to NB.  We call the corresponding classifier, lazy cell-weighted
naive Bayes (LCWNB) classifier.  

\subsection{Parametric structure of the LCWNB}
\label{sec4.4}
For any two realizations $x=(x_1, \ldots, x_m)$ and $x^*=(x_1^*, \ldots, x_m^*)$ of $X$, the Hamming distance of $x$ and $x^*$ is the total number of $i$'s ($i =1, \ldots, m$) such that $x_i \neq x_i^*$.  Denote the Hamming distance between $x$ and $x^*$ by $H(x,x^*)$.
For each $Y$-value, $y$, choose a
constant $\gamma_{y}$ such that $0 \leq \gamma_{y} \leq 1$.  
Given a test instance $x^*$, we attach to cell $(x,y)$ a weight $w_{x^*}(x,y)= \gamma_{y}^{H(x,x^*)}$. 
(We use the convention that $0^{0}=1$).  
As the cell weight is a non-increasing function of the Hamming distance, $w_{x^*}(x,y)$ is a local weighting function. 

When $\gamma_{y}=0$ for all $y$, 
only the cells $(x^{*},y)$
for different $y$ have weight one and all other cells have weight zero. 
The estimator $\hat y_w^*$ in Section 2 is the Bayesian classification rule for the full multinomial model. Therefore, the magnitude of $\gamma_{y}$
determines how much we move from NB when all $\gamma_{y}$'s are
one to the full multinomial model when all $\gamma_{y}$'s are zero. 

To prove the compatibility of the weighting function, let 
the test instance be $X= x^{*}$. 
Under CIA in $\mathcal{T}$, for any $x=(x_1, \ldots, x_m)$
\begin{eqnarray*}
& & \mbox{Pr}_{w_{x^*}}(X=x,Y=y) \\
& \propto & \Pr(X = x,Y=y) \gamma_{y}^{H(x,x^*)}\\
 & = & \Pr(Y=y) \gamma_y^{H(x,x^*)} \prod_{i=1}^{m}\Pr(X_{i}=x_{i}|Y=y).
\end{eqnarray*}
Therefore,
\begin{eqnarray*}
& & \mbox{Pr}_{w_{x^*}}(Y=y) \\
& \propto & \Pr(Y=y)\prod_{i=1}^{m}\left[\Pr(X_{i}=x_{i}^{*}|Y=y)+(1-\Pr(X_{i}=x_{i}^{*}|Y=y)) \gamma_{y} \right],
\end{eqnarray*}
and
\begin{eqnarray*}
 &  & \mbox{Pr}_{w_{x^*}}(X=x|Y=y)\\
 & = & \gamma_y^{H(x,x^*)} \prod_{i=1}^{m}\frac{\Pr(X_{i}=x_{i}|Y=y) }{\Pr(X_{i}=x_{i}^{*}|Y=y)+[1-\Pr(X_{i}=x_{i}^{*}|Y=y)] \gamma_{y}}.
\end{eqnarray*}
Clearly Equation (1) holds and Condition (a) follows.  The correctness of Condition (b) is obvious.

\subsection{Parameter selection in the LCWNB}
\label{sec4.2}
The choice of $\gamma_{y}$ has dominating effect on the new method. It controls how much information
in $\mathcal{T}$ is retained for classification. 
A simple way to quantify the retained
information is to count the number of random sample that can be generated
from $(X,Y)_{w_{x^*}}$ using $\mathcal{T}$. Such
a random sample can be drawn using the following acceptance-rejection
method. For each training instance $(x,y)$
in $\mathcal{T}$, include this instance in a set $B$ with probability
$\gamma_y^{H(x,x^*)}$. Then the training instances in $B$ form
a random sample of $(X,Y)_{w_{x^*}}$. 
Let $S_{y}$ be the expected
frequency of $Y=y$ in $B$, and $V_\ell(y)$ be the total number of training instances in $\mathcal{T}$ with $H(x,x^*) = \ell$ and $Y=y$. Then 
\[
S_{y}=\sum_{\ell=0}^m V_\ell(y) \gamma_y^{\ell}.
\]
For the $k$-nearest neighbor ($k$-NN) method where only the $k$ nearest
neighbors of the test instance have weight one and all other instances have weight zero, a similar acceptance-rejection method yields $S_y = k$.
Thus $S_{y}$ acts like the constant $k$ in the
$k$-NN method. Its value should be chosen to balance the
bias and the variability. 

Let $r$ be the number of classes.  For NB, the training set
can be partitioned into $r$ independent random samples,
one for each class label.  
It is desirable to control the degree
of localization separately for each sample. For $y$ with large frequency,
we can afford using small $\gamma_{y}$ to emphasize 
model fitting in a small region. However, for $y$ with small frequency,
large $\gamma_{y}$ is preferred as we want to retain enough information
for estimation. Assigning different $\gamma_{y}$ value for different $y$ is a means to handle class imbalance. The following simple
rule is proposed. 

\vspace{0.3cm}
\noindent 
\textbf{Simple selection rule}: Choose a positive real number $\kappa$.
Select $\gamma_{y}$ so that the corresponding $S_{y}$ is closest to $\kappa$, i.e. $S_y =\min(\max(V_0(y),\kappa),\sum_{\ell=0}^m V_\ell(y))$.

\vspace{0.3cm}
\noindent 
As $S_{y}$
is a monotonically increasing polynomial of $\gamma_y \in [0,1]$,
the value of $\gamma_{y}$ 
can be efficiently found using binary search. As an analogy of $k$
in the $k$-NN method, $\kappa$ should be small. In the comparison study, $\kappa$ is selected to be 5, 10 or 20. 

A desirable property of the simple selection rule is that 
when the number of test instances increases,
all $\gamma_{y}$'s will eventually be zero, and 
the classifier approaches the Bayesian classification rule for the full multinomial model.

\subsection{Laplace's estimator of probabilities}
\label{sec4.3}
The probability estimators in Section 2 are unreliable when zero or very small cell frequencies are encountered.
Laplace's law
of succession \citep{cestnik1990} is a common remedy of the problem.

For the unweighted case, $\gamma_y = 1$ for all $y$.
The total weight is $n$, which is the size of $\mathcal{T}$. 
As Laplace's estimator is designed for the unweighted data, it
is desirable to define a ``sample size'' for a weighted sample.
For importance estimator, a corresponding measure is 
the effective sample size. An effective sample size of a weighted average
(weighted relative frequency is a special kind of weighted average)
is $u$ if this importance estimator has the same variance as the simple
average of $u$ random sample from the target distribution. 
An approximate effective sample size
for importance estimator \citep[see][Section 2.5.3]{liu2001} 
(note that their weights are scaled so that the mean weight is 1) is
\[
\hat{C} = \frac{\left(\sum_{k} \sum_{\ell=0}^m V_\ell(k) \gamma_k^\ell \right)^2 }{\sum_{k} \sum_{\ell=0}^m V_\ell(k) \gamma_k^{2 \ell}}.
\]
To rescale the weights so that the total revised weight is equal to $\hat C$, we multiply each weight 
by a constant $\rho$, where
\[
\rho = \frac{\sum_{k} \sum_{\ell=0}^m V_\ell(k) \gamma_k^\ell }{\sum_{k} \sum_{\ell=0}^m V_\ell(k) \gamma_k^{2 \ell}} \geq 1.
\]
The Laplace's probability estimators for the rescaled weights are
\[
\widehat{\Pr}_{w_{x^*}}(Y=y)=\frac{\alpha +\rho \sum_{\ell=0}^m V_\ell(y) \gamma_y^\ell
}{\beta + \rho \sum_{k} \sum_{\ell=0}^m V_\ell(k) \gamma_k^\ell}
\]
and
\begin{eqnarray*}
& & \widehat{\Pr}_{w_{x^*}}(X_{i} = x_{i}^{*}|Y=y) \\
& = & \frac{\zeta_{i}+\rho \sum_{x_{1},\ldots x_{i-1},x_{i+1},\ldots,x_{m}}f((x_{1},...,x_{i}^{*},...,x_{m}),y) \gamma_y^{H((x_{1},...,x_{i}^{*},...,x_{m}),x^*)}}{\tau_{i}+\rho \sum_{\ell=0}^m V_\ell(y) \gamma_y^\ell}.
\end{eqnarray*}
In this paper, we use a common choice of the constants which are $\alpha=1$,
$\beta=r$ (the total number of possible class labels), $\zeta_{i}=1$ and $\tau_{i}=q_{i}$ which is the number of possible values of $X_{i}$.

LWNB \citep{frank2003} also rescales their weights.  
For LWNB, the weights are multiplied by a constant to make the total weight equal to the total number of
non-zero weights.  Their method cannot be applied to our classifier as we give all training instances positive weight.  We need 
another measure of ``sample size'' that takes into account the distribution of weights. A common feature of his method with ours is that
both multiplication factors are larger than 1.

\subsection{Implementation details}
\label{sec:Implimentation-details}
The pseudo-code of the LCWNB classifier is given in Figure \ref{tab:1}.
The time complexity for classifying one test instance is $O(mn+rmK)$,
where $n$ is the total number of training instances, $r$ is the
number of classes, and $K$ is the number of iterations in the binary
search for the parameter $\gamma_{y}$ (the difference between the computed
value and the true value of $\gamma_{y}$ is bounded by $2^{-K}$).
The term $mn$ is usually dominating as a large value of $K$ is
unnecessary. As we need to
store the whole training data set, the memory complexity is $O(mn)$.

\begin{figure}
\begin{tabular}{l}
\hline
\begin{minipage}[t]{11cm}%
\begingroup\linespread{0.8}
\vspace{0.05cm}
\begin{description}
\item [{{\bf Algorithm}}] {LazyCellWeightedNB$\left(\kappa,\mathcal{T},x^*=\left(x_{1}^{*},\ldots,x_{m}^{*}\right)\right)$}{\par}
\item [{{\bf Input:}}] {The target expected sample size $\kappa$, the
training data set $\mathcal{T}=\left\{(x_{(j)}, y_j) \right\}_{j=1, \ldots, n}$ where $x_{(j)} = (x_{j,1},...,x_{j,m} )$,
and a test instance $x^* = \left(x_{1}^{*},\ldots,x_{m}^{*}\right)$}{\small \par}
\item [{{\bf Output:}}] {The estimated class $\hat{y}$}{\par}
\end{description}
\begin{enumerate}

\item {Find $V_0(y), \ldots, V_m(y)$ for all class $y$.} { \par}

\item {For each class $y$, 
perform binary search for $\gamma_{y}\in[0,1]$ so that $\sum_{\ell=0}^m V_\ell(y) \gamma_y^\ell$ is closest to $\kappa$.
Let $S_y = \sum_{\ell=0}^m V_\ell(y) \gamma_y^\ell$ for the selected $\gamma_y$. }{\par}

\item {Compute
\[
T(i,y) = \sum_{j=1}^n \gamma_y^{H(x_{(j)},x^*)} \delta(x_{j,i},x_i^*) \delta(y_j,y)
\]
for all class $y$ and $i =1, \ldots, m$, where $\delta(\cdot , \cdot)=1$
if its two arguments are equal and is equal to 0 otherwise. 
}{\par}

\item {Calculate the multiplier of weight
\[
\rho = \frac{\sum_{k} S_k}{\sum_{k} \sum_{\ell=0}^m V_\ell(k) \gamma_k^{2 \ell}}.
\]}{\par}

\item {For each class $y$, compute
\[
Q(y)=\left( 1+ \rho S_y \right) \prod_{i=1}^{m}\frac{1+ \rho T(i,y)}{q_{i}+ \rho S_y}.
\]
}{\par}

\item {Return
\[
\hat{y}= \mbox{arg max}_{y}Q(y).
\]
\vspace{0.05cm}
}{\par}
\end{enumerate}
\endgroup%
\end{minipage}
\\
\hline
{\par}
\end{tabular}
\caption{Pseudo-code of the LCWNB classifier}
\label{tab:1}
\end{figure}

\section{Empirical comparison}
\label{sec:Empirical-comparison}
In this section, we conduct an empirical study of LCWNB.
The aim is twofold. First, we look for a good choice of the parameter
$\kappa$ for the LCWNB method. Three candidate values of $\kappa$,
namely 5, 10 and 20, are considered.
The LCWNB with these three values of $\kappa$ are denoted as LCWNB5,
LCWNB10 and LCWNB20 respectively. Second, we compare LCWNB with an appropriate
choice of $\kappa$ with seven existing methods. They are
(i) NB: naive-Bayes classifier, (ii) TAN: the tree augmented naive-Bayes
(TAN) \citep{friedman1996,friedman1997b}, (iii) AODE: the averaged
one-dependence estimator \citep{webb2005}, (iv) WAODE: the weighted
average one-dependence estimator \citep{cerquides2005}, (v) HNB:
the hidden naive-Bayes method \citep{zhang2005}, (vi) LWNB: the locally
weighted naive Bayes method \citep{frank2003}, and (vii) ICLNB: the
instance cloning local naive Bayes \citep{jiang2005}.

A collection of 36 benchmark data sets from the UCI repository \citep{frank2010} are downloaded
from the website of Weka \citep{witten2011}. They are used as test beds for the classifiers.  Summary description of the data sets is given
in Table \ref{table:datasets}. 

\begin{table}
\caption{\label{table:datasets}Summary description for classification data sets ($n$: number of training instances; $m$: number of attributes; $r$ number of class labels)}

\centering{}{\small }%
\begin{tabular}{lrrrclrrr}
\hline 
{\small Datasets} & {\small $n$} & {\small $m$} & {\small $r$} & ~~~~~ & {\small Datasets} & {\small $n$} & {\small $m$} & {\small $r$}\tabularnewline
\hline 
{\small anneal } & {\small 898} & {\small 39} & {\small 6} & & {\small ionosphere } & {\small 351} & {\small 35} & {\small 2}\tabularnewline
{\small anneal.ORIG } & {\small 898} & {\small 39} & {\small 6} & & {\small iris } & {\small 150} & {\small 5} & {\small 3}\tabularnewline
{\small audiology } & {\small 226} & {\small 70} & {\small 24} & & {\small kr-vs-kp } & {\small 3196} & {\small 37} & {\small 2}\tabularnewline
{\small autos } & {\small 205} & {\small 26} & {\small 7} & & {\small labor } & {\small 57} & {\small 17} & {\small 2}\tabularnewline
{\small balance-scale } & {\small 625} & {\small 5} & {\small 3} & & {\small letter } & {\small 20000} & {\small 17} & {\small 26}\tabularnewline
{\small breast-cancer } & {\small 286} & {\small 10} & {\small 2} & & {\small lymph} & {\small 148} & {\small 19} & {\small 4}\tabularnewline
{\small breast-w} & {\small 699} & {\small 10} & {\small 2} & & {\small mushroom } & {\small 8124} & {\small 23} & {\small 2}\tabularnewline
{\small colic } & {\small 368} & {\small 23} & {\small 2} & & {\small primary-tumor } & {\small 339} & {\small 18} & {\small 21}\tabularnewline
{\small colic.ORIG } & {\small 368} & {\small 28} & {\small 2} & & {\small segment } & {\small 2310} & {\small 20} & {\small 7}\tabularnewline
{\small credit-a} & {\small 690} & {\small 16} & {\small 2} & & {\small sick } & {\small 3772} & {\small 30} & {\small 2}\tabularnewline
{\small credit -g} & {\small 1000} & {\small 21} & {\small 2} & & {\small sonar } & {\small 208} & {\small 61} & {\small 2}\tabularnewline
{\small diabetes } & {\small 768} & {\small 9} & {\small 2} & & {\small soybean } & {\small 683} & {\small 36} & {\small 19}\tabularnewline
{\small glass } & {\small 214} & {\small 10} & {\small 7} & & {\small splice } & {\small 3190} & {\small 62} & {\small 3}\tabularnewline
{\small heart-c} & {\small 303} & {\small 14} & {\small 5} & & {\small vehicle } & {\small 846} & {\small 19} & {\small 4}\tabularnewline
{\small heart-h} & {\small 294} & {\small 14} & {\small 5} & & {\small vote } & {\small 435} & {\small 17} & {\small 2}\tabularnewline
{\small heart-statlog } & {\small 270} & {\small 14} & {\small 2} & & {\small vowel } & {\small 990} & {\small 14} & {\small 11}\tabularnewline
{\small hepatitis } & {\small 155} & {\small 20} & {\small 2} &  & {\small waveform-5000} & {\small 5000} & {\small 41} & {\small 3}\tabularnewline
{\small hypothyroid } & {\small 3772} & {\small 30} & {\small 4} & & {\small zoo } & {\small 101} & {\small 18} & {\small 7}\tabularnewline
\hline 
\end{tabular}
\end{table}

Classification accuracy rate is used as a performance measure in this
paper. The rates are computed using 10 independent runs of 10-fold
cross-validation. All classifiers are trained and tested
on exactly the same cross validation folds. In the study, the filter
ReplaceMissingValues in Weka is used to replace the missing values,
and then the filter Discretization Weka is used to perform unsupervised
10-bin discretization. If the number of values of an attribute is
almost equal to the number of instances, that attribute is removed
from the data in the preprocessing step.

Table \ref{table:exres} lists the classification accuracy rates
of the methods when applied to the data sets. Some other statistics
are given in the bottom three rows. 

\begin{table}[p]
\centering
\rotatebox{90}{
\begingroup\tabcolsep=0pt\def\arraystretch{0.7}
\begin{minipage}{\textheight}
\caption{\label{table:exres}Experimental results: percentage of correct classifications} 
\begin{tabular}[t]{lllllllllll}
\hline 
{\tiny Datasets} & {\tiny LCWNB5} & {\tiny LCWNB10} & {\tiny LCWNB20} & {\tiny NB} & {\tiny TAN} & {\tiny AODE} & {\tiny WAODE} & {\tiny HNB} & {\tiny LWNB} & {\tiny ICLNB}\tabularnewline
\hline 
{\tiny anneal } & {\tiny 98.82$\pm$1.01 } & {\tiny 98.68$\pm$1.08 } & {\tiny 98.50$\pm$1.12 } & {\tiny 94.32$\pm$2.23 $\bullet$ } & {\tiny 98.34$\pm$1.18 } & {\tiny 96.83$\pm$1.66 $\bullet$ } & {\tiny 98.56$\pm$1.22 } & {\tiny 98.62$\pm$1.14 } & {\tiny 98.41$\pm$1.14 } & {\tiny 98.82$\pm$1.06 }\tabularnewline
{\tiny anneal.ORIG } & {\tiny 93.07$\pm$2.28 } & {\tiny 92.82$\pm$2.38 } & {\tiny 92.33$\pm$2.38 } & {\tiny 88.16$\pm$3.06 $\bullet$ } & {\tiny 90.93$\pm$2.53 $\bullet$ } & {\tiny 89.01$\pm$3.10 $\bullet$ } & {\tiny 89.80$\pm$2.99 $\bullet$ } & {\tiny 91.60$\pm$2.63 $\bullet$ } & {\tiny 89.87$\pm$2.60 $\bullet$ } & {\tiny 91.71$\pm$2.29 $\bullet$ }\tabularnewline
{\tiny audiology } & {\tiny 77.35$\pm$6.26 } & {\tiny 77.18$\pm$5.45 } & {\tiny 75.14$\pm$5.88 } & {\tiny 71.40$\pm$6.37 $\bullet$ } & {\tiny 72.63$\pm$7.06 } & {\tiny 71.66$\pm$6.42 $\bullet$ } & {\tiny 76.26$\pm$6.36 } & {\tiny 73.15$\pm$6.00 } & {\tiny 74.00$\pm$6.89 } & {\tiny 78.16$\pm$7.46 }\tabularnewline
{\tiny autos } & {\tiny 76.99$\pm$9.49 } & {\tiny 74.27$\pm$10.42 } & {\tiny 69.66$\pm$10.23 $\bullet$ } & {\tiny 63.97$\pm$11.35 $\bullet$ } & {\tiny 76.97$\pm$9.16 } & {\tiny 74.60$\pm$10.10 } & {\tiny 80.36$\pm$9.48 } & {\tiny 78.04$\pm$9.43 } & {\tiny 77.50$\pm$9.74 } & {\tiny 80.00$\pm$9.09 }\tabularnewline
{\tiny balance-scale } & {\tiny 87.41$\pm$2.37 } & {\tiny 88.85$\pm$2.24 $\circ$ } & {\tiny 90.05$\pm$1.81 $\circ$ } & {\tiny 91.44$\pm$1.30 $\circ$ } & {\tiny 86.22$\pm$2.82 } & {\tiny 89.78$\pm$1.88 $\circ$ } & {\tiny 89.28$\pm$2.12 $\circ$ } & {\tiny 89.65$\pm$2.42 $\circ$ } & {\tiny 84.64$\pm$2.93 $\bullet$ } & {\tiny 84.77$\pm$2.95 $\bullet$ }\tabularnewline
{\tiny breast-cancer } & {\tiny 71.58$\pm$8.07 } & {\tiny 72.56$\pm$7.38 } & {\tiny 72.56$\pm$7.36 } & {\tiny 72.94$\pm$7.71 } & {\tiny 70.09$\pm$7.68 } & {\tiny 72.73$\pm$7.01 } & {\tiny 71.97$\pm$6.79 } & {\tiny 70.23$\pm$6.49 } & {\tiny 74.63$\pm$5.19 } & {\tiny 71.56$\pm$5.97 }\tabularnewline
{\tiny breast-w} & {\tiny 97.37$\pm$1.73 } & {\tiny 97.44$\pm$1.68 } & {\tiny 97.44$\pm$1.68 } & {\tiny 97.30$\pm$1.75 } & {\tiny 94.91$\pm$2.37 $\bullet$ } & {\tiny 96.85$\pm$1.90 } & {\tiny 96.57$\pm$2.22 } & {\tiny 96.08$\pm$2.46 } & {\tiny 96.42$\pm$2.20 } & {\tiny 97.10$\pm$1.89 }\tabularnewline
{\tiny colic } & {\tiny 81.99$\pm$6.02 } & {\tiny 81.45$\pm$6.27 } & {\tiny 80.93$\pm$6.36 } & {\tiny 78.86$\pm$6.05 $\bullet$ } & {\tiny 80.57$\pm$5.90 } & {\tiny 80.93$\pm$6.16 } & {\tiny 80.66$\pm$6.58 } & {\tiny 81.25$\pm$6.27 } & {\tiny 81.16$\pm$6.10 } & {\tiny 76.99$\pm$6.99 $\bullet$ }\tabularnewline
{\tiny colic.ORIG } & {\tiny 76.88$\pm$6.87 } & {\tiny 77.09$\pm$6.59 } & {\tiny 76.52$\pm$6.81 } & {\tiny 74.21$\pm$7.09 } & {\tiny 76.06$\pm$6.01 } & {\tiny 75.38$\pm$6.41 } & {\tiny 75.93$\pm$6.69 } & {\tiny 75.50$\pm$6.57 } & {\tiny 75.36$\pm$5.38 } & {\tiny 74.68$\pm$6.51 }\tabularnewline
{\tiny credit-a} & {\tiny 86.52$\pm$3.83 } & {\tiny 86.67$\pm$3.82 } & {\tiny 86.30$\pm$3.64 } & {\tiny 84.74$\pm$3.83 } & {\tiny 84.41$\pm$4.48 $\bullet$ } & {\tiny 85.86$\pm$3.72 } & {\tiny 84.43$\pm$3.86 $\bullet$ } & {\tiny 84.84$\pm$4.43 } & {\tiny 86.39$\pm$4.05 } & {\tiny 84.88$\pm$4.23 }\tabularnewline
{\tiny credit -g} & {\tiny 75.42$\pm$3.54 } & {\tiny 76.33$\pm$3.53 } & {\tiny 76.78$\pm$3.51 $\circ$ } & {\tiny 75.93$\pm$3.87 } & {\tiny 75.86$\pm$3.58 } & {\tiny 76.45$\pm$3.88 } & {\tiny 76.38$\pm$3.78 } & {\tiny 76.86$\pm$3.64 } & {\tiny 73.61$\pm$2.77 } & {\tiny 73.48$\pm$3.07 }\tabularnewline
{\tiny diabetes } & {\tiny 74.91$\pm$4.62 } & {\tiny 75.32$\pm$4.42 } & {\tiny 75.47$\pm$4.49 } & {\tiny 75.68$\pm$4.85 } & {\tiny 75.09$\pm$4.96 } & {\tiny 76.57$\pm$4.53 } & {\tiny 75.83$\pm$4.80 } & {\tiny 75.83$\pm$4.86 } & {\tiny 71.98$\pm$3.83 $\bullet$ } & {\tiny 73.81$\pm$4.17 }\tabularnewline
{\tiny glass } & {\tiny 63.92$\pm$9.07 } & {\tiny 64.33$\pm$9.87 } & {\tiny 61.92$\pm$9.28 } & {\tiny 57.69$\pm$10.07 $\bullet$ } & {\tiny 58.43$\pm$8.86 } & {\tiny 61.73$\pm$9.69 } & {\tiny 59.62$\pm$9.40 } & {\tiny 59.33$\pm$8.83 $\bullet$ } & {\tiny 65.46$\pm$8.53 } & {\tiny 66.40$\pm$9.03 }\tabularnewline
{\tiny heart-c} & {\tiny 81.62$\pm$7.03 } & {\tiny 82.09$\pm$7.07 } & {\tiny 82.45$\pm$6.71 } & {\tiny 83.44$\pm$6.27 } & {\tiny 82.85$\pm$7.20 } & {\tiny 82.84$\pm$7.03 } & {\tiny 82.61$\pm$7.19 } & {\tiny 81.43$\pm$7.35 } & {\tiny 81.51$\pm$7.05 } & {\tiny 78.65$\pm$8.04 }\tabularnewline
{\tiny heart-h} & {\tiny 82.70$\pm$5.89 } & {\tiny 82.73$\pm$5.84 } & {\tiny 83.38$\pm$6.00 } & {\tiny 83.64$\pm$5.85 } & {\tiny 82.14$\pm$6.20 } & {\tiny 84.09$\pm$6.00 } & {\tiny 83.11$\pm$5.79 } & {\tiny 80.72$\pm$6.00 } & {\tiny 83.41$\pm$5.83 } & {\tiny 80.96$\pm$6.15 }\tabularnewline
{\tiny heart-statlog } & {\tiny 82.19$\pm$5.85 } & {\tiny 82.15$\pm$5.70 } & {\tiny 82.56$\pm$6.03 } & {\tiny 83.78$\pm$5.41 } & {\tiny 79.37$\pm$6.87 } & {\tiny 83.63$\pm$5.32 } & {\tiny 82.30$\pm$5.66 } & {\tiny 81.74$\pm$5.94 } & {\tiny 81.81$\pm$5.70 } & {\tiny 79.78$\pm$6.19 }\tabularnewline
{\tiny hepatitis } & {\tiny 84.31$\pm$9.53 } & {\tiny 84.37$\pm$9.36 } & {\tiny 84.57$\pm$9.29 } & {\tiny 84.06$\pm$9.91 } & {\tiny 82.40$\pm$8.68 } & {\tiny 85.21$\pm$9.36 } & {\tiny 84.14$\pm$9.22 } & {\tiny 82.71$\pm$9.95 } & {\tiny 85.61$\pm$7.22 } & {\tiny 84.29$\pm$7.70 }\tabularnewline
{\tiny hypothyroid } & {\tiny 93.09$\pm$0.64 } & {\tiny 93.32$\pm$0.64 $\circ$ } & {\tiny 93.50$\pm$0.58 $\circ$ } & {\tiny 92.79$\pm$0.73 } & {\tiny 93.23$\pm$0.68 } & {\tiny 93.56$\pm$0.61 $\circ$ } & {\tiny 93.54$\pm$0.57 $\circ$ } & {\tiny 93.28$\pm$0.52 } & {\tiny 90.90$\pm$1.28 $\bullet$ } & {\tiny 93.23$\pm$0.59 }\tabularnewline
{\tiny ionosphere } & {\tiny 91.74$\pm$4.33 } & {\tiny 91.34$\pm$4.41 } & {\tiny 91.20$\pm$4.43 } & {\tiny 90.86$\pm$4.33 } & {\tiny 92.23$\pm$4.36 } & {\tiny 91.74$\pm$4.28 } & {\tiny 92.94$\pm$3.80 } & {\tiny 93.02$\pm$3.98 } & {\tiny 92.34$\pm$4.20 } & {\tiny 91.97$\pm$4.58 }\tabularnewline
{\tiny iris } & {\tiny 94.73$\pm$5.86 } & {\tiny 94.87$\pm$6.06 } & {\tiny 95.40$\pm$5.97 } & {\tiny 94.33$\pm$6.79 } & {\tiny 91.67$\pm$7.18 } & {\tiny 94.00$\pm$5.88 } & {\tiny 95.73$\pm$4.79 } & {\tiny 93.93$\pm$6.00 } & {\tiny 93.93$\pm$6.07 } & {\tiny 94.80$\pm$6.33 }\tabularnewline
{\tiny kr-vs-kp } & {\tiny 97.72$\pm$0.81 } & {\tiny 97.24$\pm$0.88 $\bullet$ } & {\tiny 96.80$\pm$0.90 $\bullet$ } & {\tiny 87.79$\pm$1.91 $\bullet$ } & {\tiny 92.05$\pm$1.49 $\bullet$ } & {\tiny 91.03$\pm$1.66 $\bullet$ } & {\tiny 94.18$\pm$1.25 $\bullet$ } & {\tiny 92.35$\pm$1.32 $\bullet$ } & {\tiny 97.26$\pm$0.86 $\bullet$ } & {\tiny 97.74$\pm$0.75 }\tabularnewline
{\tiny labor } & {\tiny 94.37$\pm$10.09 } & {\tiny 94.90$\pm$9.61 } & {\tiny 96.50$\pm$7.45 } & {\tiny 96.70$\pm$7.27 } & {\tiny 90.33$\pm$10.96 } & {\tiny 94.57$\pm$9.72 } & {\tiny 91.73$\pm$12.06 } & {\tiny 90.87$\pm$13.15 } & {\tiny 94.57$\pm$10.28 } & {\tiny 93.17$\pm$11.16 }\tabularnewline
{\tiny letter } & {\tiny 90.95$\pm$0.59 } & {\tiny 89.28$\pm$0.61 $\bullet$ } & {\tiny 86.83$\pm$0.71 $\bullet$ } & {\tiny 70.09$\pm$0.93 $\bullet$ } & {\tiny 83.11$\pm$0.75 $\bullet$ } & {\tiny 85.54$\pm$0.68 $\bullet$ } & {\tiny 88.86$\pm$0.55 $\bullet$ } & {\tiny 86.13$\pm$0.69 $\bullet$ } & {\tiny 91.47$\pm$0.48 $\circ$ } & {\tiny 92.87$\pm$0.40 $\circ$ }\tabularnewline
{\tiny lymph} & {\tiny 87.59$\pm$8.61 } & {\tiny 87.11$\pm$8.81 } & {\tiny 87.11$\pm$8.66 } & {\tiny 85.97$\pm$8.88 } & {\tiny 84.07$\pm$8.93 } & {\tiny 85.46$\pm$9.32 } & {\tiny 84.16$\pm$8.74 } & {\tiny 82.93$\pm$8.96 $\bullet$ } & {\tiny 85.03$\pm$8.82 } & {\tiny 83.81$\pm$8.67 }\tabularnewline
{\tiny mushroom } & {\tiny 100.00$\pm$0.00 } & {\tiny 100.00$\pm$0.00 } & {\tiny 100.00$\pm$0.00 } & {\tiny 95.52$\pm$0.78 $\bullet$ } & {\tiny 99.99$\pm$0.03 } & {\tiny 99.95$\pm$0.07 $\bullet$ } & {\tiny 99.98$\pm$0.04 } & {\tiny 99.96$\pm$0.06 } & {\tiny 100.00$\pm$0.00 } & {\tiny 100.00$\pm$0.00 }\tabularnewline
{\tiny primary-tumor } & {\tiny 47.02$\pm$6.13 } & {\tiny 47.94$\pm$5.62 } & {\tiny 48.38$\pm$6.05 } & {\tiny 47.20$\pm$6.02 } & {\tiny 46.76$\pm$5.92 } & {\tiny 47.87$\pm$6.37 } & {\tiny 47.94$\pm$5.89 } & {\tiny 47.85$\pm$6.06 } & {\tiny 47.08$\pm$5.48 } & {\tiny 43.36$\pm$5.99 $\bullet$ }\tabularnewline
{\tiny segment } & {\tiny 95.53$\pm$1.30 } & {\tiny 94.64$\pm$1.42 $\bullet$ } & {\tiny 93.46$\pm$1.58 $\bullet$ } & {\tiny 89.03$\pm$1.66 $\bullet$ } & {\tiny 94.54$\pm$1.60 $\bullet$ } & {\tiny 92.92$\pm$1.40 $\bullet$ } & {\tiny 95.06$\pm$1.39 } & {\tiny 94.72$\pm$1.42 } & {\tiny 93.48$\pm$1.50 $\bullet$ } & {\tiny 95.57$\pm$1.15 }\tabularnewline
{\tiny sick } & {\tiny 98.25$\pm$0.66 } & {\tiny 98.17$\pm$0.66 } & {\tiny 98.05$\pm$0.66 } & {\tiny 96.78$\pm$0.91 $\bullet$ } & {\tiny 97.61$\pm$0.73 $\bullet$ } & {\tiny 97.52$\pm$0.72 $\bullet$ } & {\tiny 97.99$\pm$0.72 } & {\tiny 97.78$\pm$0.73 $\bullet$ } & {\tiny 95.42$\pm$1.30 $\bullet$ } & {\tiny 98.14$\pm$0.61 }\tabularnewline
{\tiny sonar } & {\tiny 80.21$\pm$8.92 } & {\tiny 79.73$\pm$8.92 } & {\tiny 78.39$\pm$9.34 } & {\tiny 76.35$\pm$9.94 } & {\tiny 73.66$\pm$10.04 } & {\tiny 79.91$\pm$9.60 } & {\tiny 78.28$\pm$8.51 } & {\tiny 80.89$\pm$8.68 } & {\tiny 82.76$\pm$7.78 } & {\tiny 82.08$\pm$7.76 }\tabularnewline
{\tiny soybean } & {\tiny 93.22$\pm$2.57 } & {\tiny 93.63$\pm$2.69 } & {\tiny 93.44$\pm$2.85 } & {\tiny 92.20$\pm$3.23 } & {\tiny 95.23$\pm$2.32 $\circ$ } & {\tiny 93.31$\pm$2.85 } & {\tiny 94.33$\pm$2.36 $\circ$ } & {\tiny 94.67$\pm$2.25 $\circ$ } & {\tiny 93.15$\pm$2.54 } & {\tiny 93.90$\pm$2.35 }\tabularnewline
{\tiny splice } & {\tiny 96.38$\pm$0.96 } & {\tiny 96.34$\pm$0.98 } & {\tiny 96.23$\pm$0.97 } & {\tiny 95.42$\pm$1.14 $\bullet$ } & {\tiny 95.39$\pm$1.16 $\bullet$ } & {\tiny 96.12$\pm$1.00 } & {\tiny 96.36$\pm$0.95 } & {\tiny 96.13$\pm$0.99 } & {\tiny 93.52$\pm$1.40 $\bullet$ } & {\tiny 91.44$\pm$1.51 $\bullet$ }\tabularnewline
{\tiny vehicle } & {\tiny 69.23$\pm$3.65 } & {\tiny 67.86$\pm$3.51 $\bullet$ } & {\tiny 66.59$\pm$3.05 $\bullet$ } & {\tiny 61.03$\pm$3.48 $\bullet$ } & {\tiny 73.71$\pm$3.48 $\circ$ } & {\tiny 71.65$\pm$3.59 $\circ$ } & {\tiny 73.09$\pm$3.56 $\circ$ } & {\tiny 73.63$\pm$3.86 $\circ$ } & {\tiny 71.09$\pm$3.60 $\circ$ } & {\tiny 71.33$\pm$3.77 }\tabularnewline
{\tiny vote } & {\tiny 95.63$\pm$3.08 } & {\tiny 94.80$\pm$3.20 } & {\tiny 93.88$\pm$3.42 } & {\tiny 90.21$\pm$3.95 $\bullet$ } & {\tiny 94.57$\pm$3.23 } & {\tiny 94.52$\pm$3.19 } & {\tiny 94.46$\pm$3.17 } & {\tiny 94.36$\pm$3.20 } & {\tiny 95.10$\pm$3.07 } & {\tiny 95.74$\pm$2.71 }\tabularnewline
{\tiny vowel } & {\tiny 92.27$\pm$2.63 } & {\tiny 90.01$\pm$3.17 $\bullet$ } & {\tiny 86.32$\pm$3.44 $\bullet$ } & {\tiny 66.09$\pm$4.78 $\bullet$ } & {\tiny 93.10$\pm$2.85 } & {\tiny 89.64$\pm$3.06 $\bullet$ } & {\tiny 92.56$\pm$2.70 } & {\tiny 92.99$\pm$2.49 } & {\tiny 93.78$\pm$2.54 } & {\tiny 93.69$\pm$2.39 }\tabularnewline
{\tiny waveform-5000} & {\tiny 82.24$\pm$1.52 } & {\tiny 81.99$\pm$1.50 $\bullet$ } & {\tiny 81.71$\pm$1.48 $\bullet$ } & {\tiny 79.97$\pm$1.46 $\bullet$ } & {\tiny 80.72$\pm$1.78 $\bullet$ } & {\tiny 84.24$\pm$1.60 $\circ$ } & {\tiny 84.00$\pm$1.60 $\circ$ } & {\tiny 83.58$\pm$1.61 $\circ$ } & {\tiny 78.12$\pm$1.79 $\bullet$ } & {\tiny 76.56$\pm$1.67 $\bullet$ }\tabularnewline
{\tiny zoo } & {\tiny 94.76$\pm$6.51 } & {\tiny 94.57$\pm$6.50 } & {\tiny 94.57$\pm$6.50 } & {\tiny 94.37$\pm$6.79 } & {\tiny 96.73$\pm$5.45 } & {\tiny 94.66$\pm$6.38 } & {\tiny 98.11$\pm$3.92 } & {\tiny 99.90$\pm$1.00 $\circ$ } & {\tiny 96.25$\pm$5.41 } & {\tiny 97.73$\pm$4.42 }\tabularnewline
\hline 
{\tiny Average } & {\tiny 85.78 } & {\tiny 85.61 } & {\tiny 85.14 } & {\tiny 82.34 } & {\tiny 84.33 } & {\tiny 85.07 } & {\tiny 85.59 } & {\tiny 85.18 } & {\tiny 85.20 } & {\tiny 85.09 }\tabularnewline
\hline 
{\tiny Mean rank} & {\tiny 4.4167} & {\tiny 4.3194} & {\tiny 4.8472} & {\tiny 7.3611} & {\tiny 7.0417} & {\tiny 5.4583} & {\tiny 4.7500} & {\tiny 5.6389} & {\tiny 5.6944} & {\tiny 5.4722}\tabularnewline
\hline 
\multicolumn{2}{l}{{\tiny {[}s.w./m.w./m.b./s.b.{]} }} & {\tiny {[}2/14.5/13.5/6{]} } & {\tiny {[}3/11.5/14.5/7{]} } & {\tiny {[}1/8/11/16{]}$\bullet$ } & {\tiny {[}2/7/18/9{]}$\bullet$ } & {\tiny {[}4/10.5/12.5/9{]} } & {\tiny {[}5/11/16/4{]} } & {\tiny {[}5/8/17/6{]}$\bullet$ } & {\tiny {[}2/11.5/13.5/9{]} } & {\tiny {[}1/14/15/6{]} }\tabularnewline
\hline 
\multicolumn{11}{l}{{\tiny $\circ$, $\bullet$ statistically significant improvement
or degradation}}\tabularnewline
\end{tabular}
\end{minipage}
\endgroup
}
\end{table}

Since no single classifier outperforms others
in all data sets, statistical analyses are in need in the
comparison. 
As the data sets are not randomly
chosen, all statistical results apply only to an imaginary population
where the data sets are representative. For example for the 36
data sets, the largest size of the training data is 20000. Extrapolating
the results to data exceeding this limit is unsafe. Sixteen of the
36 data sets have $r = 2$. The results are therefore biased towards
two-class classification. Some data sets have a common source.
Such common sources have larger impact on the comparison results. 

To sketch out a general picture of the accuracy rates, two descriptive performance statistics are computed for each classifier.
One is the average accuracy rate displayed in the third row from the bottom of Table \ref{table:exres}.  
It is the most
fundamental summary measure as its interpretation does not depend on what
other classifiers are included in the study. 
The other measure is the mean rank which is listed in the second row from the bottom of the same table.   
It is the average rank of the classifier, 
with rank 1 assigned to the method having the largest rate for a data set, and rank 10 to
the method having the smallest rate. 
Mean rank is robust to extraordinary accuracy rates. 

\begin{figure}
\includegraphics[width=12cm,height=5cm]{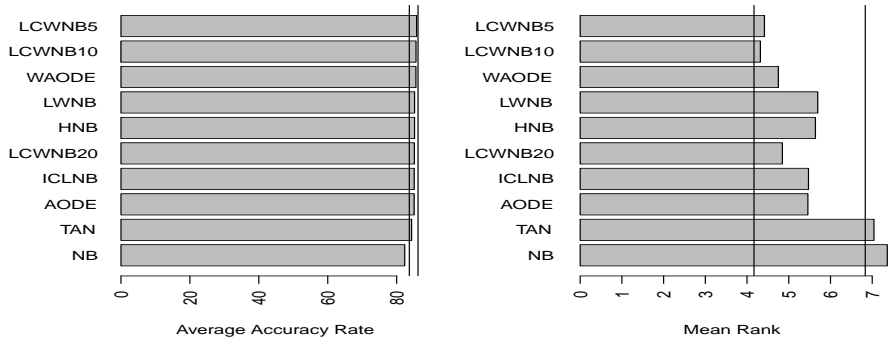}
\caption{Average accuracy rates and mean ranks for classifiers}
\label{fig:1}
\end{figure}

Figure \ref{fig:1} presents a paired bar chart for
the two measures with classifiers arranged in the descending order
of the average classification accuracy rates. The two reference lines in each bar chart are the
95\% simultaneous confidence bounds for the corresponding measure
under the assumption that all methods perform equally well. 
They are computed from 9999 random permutations of the data. As
there are bars lying outside the confidence bounds in the charts,
the equal performance hypothesis should be rejected at 5\% significance
level. NB and TAN are likely to be inferior to other methods.
LCWNB5 and LCWNB10 perform well in both measures. They are the
best two with WAODE the third best. 

The first purpose of this study is to suggest an appropriate $\kappa$-value.
Nonparametric tests are performed to compare the three $\kappa$ values.
The $p$-values for the Friedman test and the Iman and Davenport's modification \citep{iman1980} are
0.2366 and 0.2393 respectively.  The $p$-value for the Quade test \citep{quade1979} is 0.1449.
All tests indicate that the three choices are of equal performance at 5\% significance level. 
We get the same conclusion from the one-sided Wilcoxon tests 
when we investigate whether LCWNB5 is superior to LCWNB10 and LCWNB20.

When we rank these three methods, it is found that LCWNB10 is usually ranked second.  Its accuracy rate lies between
those of the other two methods in 27 out of 36 data sets.  The $p$-value for this pattern
is $\Pr(Binomial(36,1/3) \geq 27) = 3.81 \times 10^{-7}$.  
LCWNB5 is ranked 1 in 18 out of the 36 data sets.
The $p$-value for it is $\Pr(Binomial(36,1/3) \geq 18) = 0.0283$.  Both p-values suggest that   
the three methods behave differently contradicting the results of the Friedman and related tests.
This inconsistency can be explained when we discover that 
LCWNB10 is usually ranked 2 while LCWNB5 and LCWNB20 are commonly ranked 1 or 3.  As a result, 
their mean ranks are close to each other, but the distributions of ranks are different.

It is of interest to have a close inspection on the performance of
the three choices of $\kappa$ when the data set characteristics are taken into account. 
Figure \ref{fig:2} shows the ranks of the LCWNB
for the three $\kappa$-values when the data sets are arranged in ascending
order of $n$, $m$ and $r$.

\begin{figure}
{\footnotesize
\begin{verbatim}          
                                        Rank  
LCWNB5 : 3 1 1 3 3 1 1 2 1 2 3 3 3 3 1 2-1 1 3 3 2 3 3 1 1-1 1 3 1 1 1 3-1 1 2 1 
LCWNB10: 2 * * 2 2 2 2 1 2 3 * 2 2 2 2 1-2 2 2 1 1 * 2 2 2-2 2 2 2 2 2 2-2 2 2 2 
LCWNB20: 1 * * 1 1 3 3 3 3 1 * 1 1 1 3 3-3 3 1 2 3 * 1 3 3-3 3 1 3 3 3 1-3 3 2 3   
                   Data sets arranged in ascending order of n

LCWNB5 : 3-3 3 3-3-2 3-3-2-1 2 3-1-1 3-1 1-1 3-1 3 2-1 1 2 3-1 1 3 1 1-1 1 1 1 1 
LCWNB10: 2-2 2 *-*-1 2-2-3-2 1 2-2-2 2-* *-2 2-2 2 2-2 2 1 2-2 2 1 2 2-2 2 2 2 2 
LCWNB20: 1-1 1 *-*-3 1-1-1-3 3 1-3-3 1-* *-3 1-3 1 2-3 3 3 1-3 3 2 3 3-3 3 3 3 3            
                   Data sets arranged in ascending order of m

LCWNB5 : 3-3-3-3-3-3-2-2-2-2-1-1-1-1-1-1 3-3-1-1 3-1-1 3-3 1-1 2-1-1-1 1 3 3 1 1 
LCWNB10: 2-2-2-2-*-*-3-2-1-1-2-2-2-2-2-2 2-2-2-2 2-*-2 2-2 2-2 1-*-2-2 2 1 2 2 2 
LCWNB20: 1-1-1-1-*-*-1-2-3-3-3-3-3-3-3-3 1-1-3-3 1-*-3 1-1 3-3 3-*-3-3 3 2 1 3 3 
                   Data sets arranged in ascending order of r

* stands for 1.5 or 2.5
- tie with respect to the data set constants, n, m or r
\end{verbatim}
\caption{Performance of LCWNB and the data set characteristics}
\label{fig:2}
}\end{figure}

While no obvious
pattern is found in the top and the bottom panels of Figure \ref{fig:2}, the
middle panel displays a trend in the ranks. LCWNB5 is ranked 1 in the seven
data sets with the largest $m$ and ranked 3 in seven of the ten data sets with the smallest $m$.
The probability of observing as or more extreme than this discovered pattern is
$(1/3)^7 \times \Pr(Binomial(10,1/3) \geq 7) = 8.990 \times 10^{-6}$.  Similar but reversed pattern is found for
LCWNB20.  It suggests that we should use LCWNB20 when $m$
is small, and gradually reduce the $\kappa$ value when $m$ increases.
Basing on the empirical data, the optimal switching rule that minimizes
the mean rank is to use LCWNB20 when $m<15$, use LCWNB10 when $m=15$
or 16, and use LCWNB5 when $m>16$.  This close relation between $\kappa$ and $m$ is not surprising because the
smallest possible value of $w(x,y)$ is $\gamma_y^m$.  The larger the $m$, the smaller
the weight is expected.  

If we have to fix $\kappa$ to a single value, $\kappa = 5$ is a reasonable choice.
Let us use LCWNB5 as a standard and
compare it with the other nine methods. 
A paired-T-test is conducted for each data set and each of the nine
method.  Solid dots and hollow dots are added in the table to indicate whether the T-test shows a significant improvement or degradation  (when LCWNB5 is compared to the method) respectively.  The bottom row of the table summarizes the results of the 324 (= $36 \times 9$) tests 
by showing the frequencies of the following four categories: (1)
LCWNB5 is significantly worse than the method; (2) LCWNB5 is worse than
the method, but the difference is not significant; (3) LCWNB5 is better
than the method, but the difference is not significant; and (4) LCWNB5
is significantly better than the method. Tie is counted as 0.5 in
categories 2 and 3. The significance level used in the tests is 5\%.
The abbreviations, 
s.w./m.w./m.b./s.b., in the last row of Table \ref{table:exres} stand for ``significantly worse''/``marginally
worse''/``marginally better''/``significantly better.'' 
As $\Pr(Binomial(36,0.5)\geq 23)=0.0662$,
we accept the alternative hypothesis that a method is worse than LCWNB5 at 6.62\% significance
level if (m.b. + s.b.) is larger than 23. Dots are added in the table to NB, TAN
and HNB to indicate that they are significantly worse than LCWNB5 in the above test. 

Nonparametric tests for LCWNB5 and the seven existing methods are performed.  Again NB and TAN are found significantly inferior to the other methods.  We exclude NB and TAN from the study and compare LCWNB5 with the remaining methods.  
The test statistic is the mean rank of LCWNB5.
 We would accept that LCWNB5 is superior if the observed mean rank 3.0556 of LCWNB5 is too small
 to be explained by chance under the equal performance assumption.  Let $rank(i,j)$ be the rank of
 method $i$ in the $j$-th data set.  Under the null hypothesis of equal performance, the mean rank of LCWNB5  
 approximately follows the Gaussian distribution with mean 3.5 and standard deviation
 $$
 s=\frac{\sqrt{v_1 + \cdots + v_{36}}}{36},
 $$
 where $v_j = \sum_{i=1}^6 (rank(i,j) - 3.5)^2 /6$.  The (one-sided) $p$-value associated with the average rank of LCWNB5 is 
 $\Pr(N(0,1) \leq (3.0556 - 3.5)/s) = 0.05855$ showing that the mean rank of LCWNB5 is 
significantly small if the level is set at 6\%.  
 
We have discovered that it is better to choose $\kappa$ according to the number
of attributes in the data.  We denote the corresponding LCWNB method by LCWNB*
when the switching rule mentioned before is used.  Applying the same test to compare LCWNB* with
AODE, WAODE, HNB, LWNB and ICLNB, the (one-sided) $p$-value is 0.01559.  It shows that LCWNB* is significantly better than other tests at 2\% level.

\section{Discussions}
\label{sec:Discussions}
In this paper, a new locally weighted classifier is proposed.  It has close relation with the methods that use only the $k$ nearest neighbors of the test instance
\citep{xie2002,frank2003,jiang2005}. 
Their methods differ from ours in three ways. First, we control the expected sample size rather than
the number of instances with positive weight. Second, their weighting functions are not compatible to NB.
Their probability estimator is derived under an inaccurate model even when CIA holds in the
unweighted data.
Third, our weights depend on the value of $Y$ and the attributes,
while their weights depend only on the attributes.  This discrepancy can
make a significant difference when the empirical distribution of 
$Y$ is highly uneven.  Using the same weighting function for all $Y$-values can lead to unreliable
probability estimator for those $y$-values with small relative frequency.

On the whole, LCWNB is simple, and easy to understand. It is
sound theoretically and performs well empirically. It improves NB
through using probability estimator that is robust to the correctness 
of the CIA without making any additional assumption. 

\bibliographystyle{spbasic}
\bibliography{ref}

\end{document}